\documentclass[conference]{IEEEtran}
\IEEEoverridecommandlockouts
\usepackage{cite}
\usepackage{amsmath,amssymb,amsfonts}
\usepackage{algorithmic}
\usepackage{graphicx}
\usepackage{textcomp}
\usepackage{xcolor}
\usepackage{multirow}
\def\BibTeX{{\rm B\kern-.05em{\sc i\kern-.025em b}\kern-.08em
    T\kern-.1667em\lower.7ex\hbox{E}\kern-.125emX}}

\begin{document}

\title{Image Clustering using Restricted Boltzman Machine}

\author{\IEEEauthorblockN{Enoch Solomon}
\IEEEauthorblockA{\textit{Department of Computer Science} \\
Virginia State University \\
Richmond, Virginia \\
esolomon@vsu.edu
}

\and
\IEEEauthorblockN{Abraham Woubie}
\IEEEauthorblockA{\textit{Silo AI} \\
Helsinki, Finland\\
Abraham.zewoudie@silo.ai}

\and

\IEEEauthorblockN{Eyael Solomon Emiru}
\IEEEauthorblockA{\textit{Department of Information} \\
Engineering and Computer Science \\
University of Trento, Italy \\
eyael.emiru@studenti.unitn.it}
}

\maketitle

\begin{abstract}
 In various verification systems, Restricted Boltzmann Machines (RBMs) have demonstrated their efficacy in both front-end and back-end processes. In this work, we propose the use of RBMs to the image clustering tasks. RBMs are trained to convert images into image embeddings. We employ the conventional bottom-up Agglomerative Hierarchical Clustering (AHC) technique. To address the challenge of limited test face image data, we introduce Agglomerative Hierarchical Clustering based
Method for Image Clustering using Restricted
Boltzmann Machine (AHC-RBM) with two major steps. Initially, a universal RBM model is trained using all available training dataset. Subsequently, we train an adapted RBM model using the data from each test image. Finally, RBM vectors which is the embedding vector is generated by concatenating the visible-to-hidden weight matrices of these adapted models, and the bias vectors. These vectors effectively preserve class-specific information and are utilized in image clustering tasks. Our experimental results, conducted on two benchmark image datasets (MS-Celeb-1M and DeepFashion), demonstrate that our proposed approach surpasses well-known clustering algorithms such as k-means, spectral clustering, and approximate Rank-order.

\end{abstract}

\begin{IEEEkeywords}
agglomerative hierarchical clustering; image clustering; face clustering; restricted boltzman machine (RBM) adaptation; unsupervised learning
\end{IEEEkeywords}

\section{Introduction} 

Deep learning has witnessed substantial success across various domains, particularly in image and speech technologies, over recent decades. This achievement has had a profound impact on the research community, prompting the adoption of these techniques in the realm of face image recognition tasks \cite{parkhi2015deep,solomon2022uface,solomon2023face,solomon2023fass,solomon2023hdlhc,solomon2023deep,solomon2023autoencoder,solomon2023unsupervised}. In the context of face image recognition, deep learning has been employed for extracting bottleneck features (BNF) and subsequently computing Gaussian Mixture Models posterior probabilities within a hybrid Deep Neural Network–Hidden Markov Model framework \cite{chuk2017hidden}. At the forefront, deep learning exhibits its capability to acquire profound features from image data, which find utility in various image recognition tasks \cite{solomon2023face}. Deep neural networks have also been successfully used for different speech and security applications \cite{woubie2021federated,woubie2021voice,woubie2021federatedondevice,woubie2021use,cima,scope,ics_sea,memory_safety,fmemory_safety,earic,containers_security}. 

Unsupervised deep learning architectures, such as Restricted Boltzmann Machines (RBMs), Deep Belief Networks (DBNs), and Deep Autoencoders, possess the inherent potential for representation learning \cite{nair2010rectified}. An initial endeavor to incorporate RBMs into the backend of an image recognition task was introduced in a seminal work by Nair et al. \cite{nair2010rectified}. While the authors primarily focused on enhancing the front-end of the image recognition system, they utilized RBM adaptation to learn a concise image representation in the form of an embedding vector \cite{alphonse2021multi}. Additionally, DBNs have been harnessed within the PLDA framework for image recognition at the backend. The resultant vector representation of an image is referred to as an RBM embedding vector, and it has been demonstrated to effectively capture image class-specific information. This promising development has motivated our exploration of utilizing such vector representations in the context of image clustering.

\begin{figure*}[h!]
	\centering
		\includegraphics[scale=0.6]{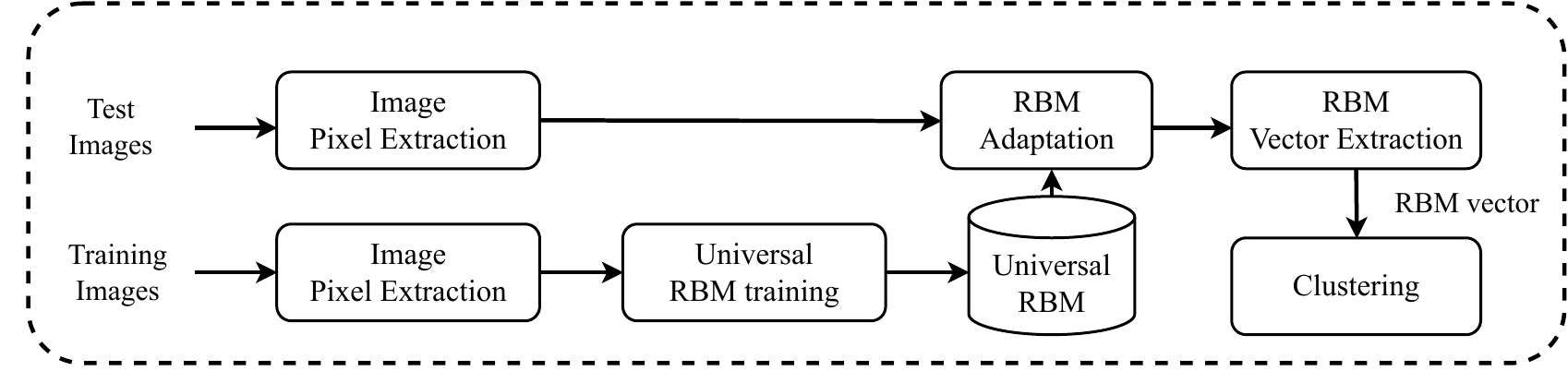}
	\caption{Block diagram showing different stages of the Restricted Boltzmann Machines vector extraction and its input to Bottom-up Agglomerative Hierarchical Clustering.}
	\label{fig:proposed}
\end{figure*}

Face image clustering is a task focused on the clustering of face images into groups, with the goal of grouping face images of the same individual within the same cluster \cite{shi2018face}. Ideally, each cluster should exclusively contain face images belonging to a single person, ensuring that face images from the same individual are not distributed across multiple clusters. Various approaches have been developed for image clustering tasks, including cost optimization, AHC and sequential methods \cite{fachrurrozi2017grouping,lin2017proximity}.

These approaches employ a range of techniques, with some relying on conventional statistical modeling like GMMs, while others leverage features extracted through DNNs. For instance, in the work by Lin et al. \cite{lin2018proximity}, bottleneck features extracted from different DNNs are utilized in a face image clustering task, employing an AHC-based approach.

To address the efficiency challenges of Graph Convolutional Networks (GCNs), STAR-FC \cite{shen2021structure} introduced a local graph learning strategy, aiming to simultaneously handle large-scale training and efficient inference. In the context of addressing noisy connections in the K-NN graph constructed in feature space, Ada-NETS \cite{wang2022ada} proposed an adaptive neighbor discovery strategy to generate cleaner graphs for GCNs.

While GCN-based methods have demonstrated significant advancements, they often utilize shallow GCNs, leading to a lack of high-order connection information. Furthermore, their efficiency remains a concern. Pair-Cls \cite{liu2021learn} proposed an alternative approach using pairwise classification instead of GCNs to reduce memory consumption and inference time. Clusformer \cite{nguyen2021clusformer} introduced an automatic visual clustering method based on Transformer \cite{vaswani2017attention}.

Various techniques have been proposed to enhance clustering performance and address computational efficiency challenges: CDP introduced an ensemble strategy to aggregate features from different models, although this approach incurred a significantly higher computational cost \cite{zhan2018consensus}. L-GCN pioneered the use of GCNs to predict linkages within an instance pivot subgraph and subsequently extracted connected components as clusters \cite{yang2019learning}. Both LTC \cite{yang2019learning} and GCN(V+E) \cite{yang2020learning} adopted a two-stage GCN approach for clustering with the K-Nearest Neighbors (K-NN) graph. LTC generated subgraphs as proposals and detected face clusters within them, while GCN(V+E) learned both confidence and connectivity using GCNs and GCN(V+E). To address the efficiency concerns associated with GCNs, STAR-FC introduced a local graph learning strategy that simultaneously tackled the challenges of large-scale training and efficient inference \cite{shen2021structure}. To mitigate noisy connections in the K-NN graph constructed in feature space, Ada-NETS proposed an adaptive neighbor discovery strategy to create clean graphs for GCNs \cite{wang2022ada}. While GCN-based methods achieved notable improvements, they primarily utilized shallow GCNs, leading to a lack of high-order connection information. Additionally, their computational efficiency remained an issue. Pair-Cls introduced the use of pairwise classification as an alternative to GCNs to reduce memory consumption and inference time \cite{liu2021learn}.Clusformer devised an automatic visual clustering method based on the Transformer architecture, which has shown significant promise in various natural language processing tasks \cite{nguyen2021clusformer} .These methods represent a range of approaches and strategies aimed at enhancing the accuracy and efficiency of face image clustering.

This paper introduces Agglomerative Hierarchical Clustering based
Method for Image Clustering using Restricted
Boltzmann Machine (AHC-RBM). The process involves several steps. Initially, a global RBM is trained using all available training data. Subsequently, an adapted RBM model is trained for each test image. In the context of face image clustering, the test face images belong to the same class that is targeted for clustering. The visible-to-hidden weight matrices, along with the visible and hidden bias vectors of these adapted RBMs, are concatenated to generate RBM supervectors which is called the RBM embedding vector.

For the face image clustering task, we employ the method outlined earlier to extract RBM vectors from the test face images. This means that all the face images intended for clustering are now represented as RBM vectors. Subsequently, we perform clustering on these embedding vectors using a bottom-up Agglomerative Hierarchical Clustering approach, considering both cosine distance and PLDA scores.

The rest of the paper is organized as follows: Section 2 provides an in-depth explanation of the process, including the training of a global model referred to as Universal RBM (URBM), RBM adaptation for face images, RBM vector extraction. Section 3 details the experimental setup, describes the database used, and outlines the evaluation metrics employed in our study. Section 4 presents the results obtained from our experiments. Finally, Section 5 offers conclusions drawn from our findings.

\section{Proposed Method}
\subsection{RBM Vector Representation}

In this  paper, we introduce a novel approach for image clustering, involving the creation of a compact, vector-based representation of images through RBM adaptation. Figure \ref{fig:proposed} provides a detailed block diagram illustrating the proposed RBM embedding vector extraction process. Initially, we train a universal model known as URBM using a substantial amount of training dataset. This URBM serves as a foundation for subsequent adaptations. The URBM is further adapted to the specific characteristics of each test image. As a result, a dedicated RBM model is trained for each test image. This adaptation process tailors the model to better represent the features within each image. To generate the desired vector representation for each image class, we utilize the visible-to-hidden weight matrices from these adapted models. These matrices play a crucial role in transforming the image data into vector representations. These embedding vector of images are subsequently utilized in the tasks mentioned earlier, leveraging cosine or probabilistic linear discriminant analysis techniques. The entire process of image embedding RBM vector extractions comprises three primary stages: URBM training, RBM adaptation, and embedding vector extraction. This approach offers a promising method for enhancing image clustering tasks by creating efficient and informative image representations.

\begin{table*}[h!]
\caption{The comparison is conducted on MS-Celeb-1M dataset by training with 0.5 million face images and testing on five distinct test subsets, each featuring varying numbers of unlabeled face images.}
\label{table:MS-Celeb-1M}
\begin{tabular}{cllllllllll}
\textbf{Images} & \multicolumn{2}{c}{\textbf{584K}}                                 & \multicolumn{2}{c}{\textbf{1.74M}}                                & \multicolumn{2}{c}{\textbf{2.89M}}                                & \multicolumn{2}{c}{\textbf{4.05M}}                                & \multicolumn{2}{c}{\textbf{5.21M}}                                \\
\textbf{Method} & \multicolumn{1}{c}{\textbf{Fp}} & \multicolumn{1}{c}{\textbf{$F_{B}$}} & \multicolumn{1}{c}{\textbf{Fp}} & \multicolumn{1}{c}{\textbf{$F_{B}$}} & \multicolumn{1}{c}{\textbf{Fp}} & \multicolumn{1}{c}{\textbf{$F_{B}$}} & \multicolumn{1}{c}{\textbf{Fp}} & \multicolumn{1}{c}{\textbf{$F_{B}$}} & \multicolumn{1}{c}{\textbf{Fp}} & \multicolumn{1}{c}{\textbf{$F_{B}$}} \\
K-means \cite{lloyd1982least}        & 79.21                           & 82.23                           & 73.04                           & 75.20                           & 69.83                           & 72.34                           & 67.90                           & 70.57                           & 66.47                           & 69.42                           \\
DBSCAN \cite{ester1996density}         & 67.93                           & 67.17                           & 63.41                           & 66.53                           & 52.50                           & 66.26                           & 45.24                           & 44.87                           & 44.94                           & 44.74                           \\
LTC \cite{yang2019learning}            & 85.66                           & 85.52                           & 82.41                           & 83.01                           & 80.32                           & 81.10                           & 78.98                           & 79.84                           & 77.87                           & 78.86                           \\
GCN(V+E) \cite{yang2020learning}      & 87.93                           & 86.09                           & 84.04                           & 82.84                           & 82.10                           & 81.24                           & 80.45                           & 80.09                           & 79.30                           & 79.25                           \\
Clusformer \cite{nguyen2021clusformer}      & 88.20                           & 87.17                           & 84.60                           & 84.05                           & 82.79                           & 82.30                           & 81.03                           & 80.51                           & 79.91                           & 79.95                           \\
STAR-FC \cite{shen2021structure}        & 91.97                           & 90.21                           & 88.28                           & 86.26                           & 86.17                           & 84.13                           & 84.70                           & 82.63                           & 83.46                           & 81.47                           \\
Ada-NETS \cite{wang2022ada}       & 92.79                           & 91.40                           & 89.33                           & 87.98                           & 87.50                           & 86.03                           & 85.40                           & 84.48                           & 83.99                           & 83.28                           \\
TPDi \cite{chen2022mitigating}      & 93.22                           & 92.18                           & 90.51                           & 89.43                           & 89.09                           & 88.00                           & 87.93                           & 86.92                           & 86.94                           & 86.06                           \\
\textbf{Ours (AHC-RBM)}   & \textbf{94.32}                  & \textbf{91.45}                  & \textbf{92.14}                  & \textbf{90.47}                  & \textbf{90.84}                  & \textbf{88.15}                  & \textbf{86.26}                  & \textbf{87.68}                  & \textbf{87.54}                  & \textbf{87.44}                 
\end{tabular}
\end{table*}

\subsubsection{URBM Training}

To obtain the embedding vector of the image, the initial step involves the training of a universal model (URBM) using an extensive dataset of available training images. URBM primary purpose is to capture class-independent information. The URBM is trained as a single model, incorporating features extracted from the entirety of the training image dataset. In the context of real-valued input features, we employ Gaussian real-valued units for the visible layer of the RBM, as proposed by Hinton et al. \cite{hinton2012practical}. The training process is carried out using the CD-1 algorithm, following the assumption that the inputs have a mean of zero and a variance of one \cite{hinton2006fast}. Consequently, the features undergo Mean Variance Normalization prior to RBM training. During this stage, the global model is trained using a substantial number of training samples generated from the feature vectors of the training images. The main objective of this global model is to learn and capture  class-specific variabilities present within the extensive training data. This approach facilitates the acquisition of a robust and class-independent representation of the input data through the URBM model.

\subsubsection{ RBM Adaptation}

Following the training of the URBM model, the next step involves image adaptation for each test image individually. The adapted RBM model is trained exclusively with data from the corresponding class of the image, allowing it to capture class-specific information. In this adaptation process, the RBM model of the class image is initialized with the parameters, including weights and biases, derived from the URBM. Essentially, the adaptation step guides the URBM model to focus on class-specific characteristics. This adaptation technique enables the RBM model to specialize for the given class. The adaptation process is carried out using the CD-1 algorithm, which ensures that the adapted RBM model aligns its parameters with the characteristics of the specific class image. Since there is only one weight matrix in an RBM, all the information learned by the RBM is encapsulated within this weight matrix. Consequently, this weight matrix is expected to convey the class-specific information relevant to the corresponding class. This approach enables the extraction of class-specific features from the adapted RBM models, enhancing the model's ability to represent and differentiate different classes of images effectively.

\subsubsection{RBM Vector Extraction}

Following the adaptation step, an RBM model is assigned to each test image. The visible-to-hidden weight matrices, along with their corresponding bias vectors, from these adapted RBMs are concatenated to create higher-dimensional image vectors. These concatenated vectors are often referred to as RBM embedding vectors. The resulting RBM vectors are expected to carry sufficient class-specific information, enabling them to effectively discriminate between different classes of images. This process enhances the quality of the image representations for subsequent tasks, such as clustering or classification.

\subsection{Image Clustering}

To assess the impact of RBM vectors on image clustering, we examined the traditional bottom-up AHC clustering method, offering both single and average linkage options. We opted not to explore the model retraining approach due to its computationally intensive nature, especially when compared to the more efficient linkage-based clustering techniques \cite{wang2019linkage,guo2020density}. The system initiates with an initial number of clusters equal to the total number of images. Through iterative steps, images that are more likely to belong to the same classes are merged until a predefined stopping condition is met. This stopping condition can either involve applying a threshold to the similarity scores to decide on cluster merging or achieving a predetermined target number of clusters. The clustering process relies on the computation of a similarity matrix, denoted as $M(X)$, among all the images in the set of identity images to be clustered, where $X$ represents this set. Consequently, RBM vectors for all images are extracted, and the matrix $M(X)$ is constructed by scoring all RBM vectors against each other. For a set of $N$ RBM vectors, the dimensions of the matrix $M(X)$ are $N \times N$. In each iteration, images with the minimum or maximum similarity scores are merged, and the matrix $M(X)$ is updated accordingly. The rows and columns corresponding to the clustered images are removed from $M(X),$ and a new row and column are introduced. These new rows and columns contain the similarity scores between the newly formed clusters and the original ones. The computation of these new scores adheres to the linkage algorithm in use. For instance, when images $S_a$ and $S_b$ are clustered into $S_{a b},$ the scores between the new cluster $\left(S_{a b}\right)$ and the original images $\left(S_n\right)$ are determined as follows:

(a) Average Linkage:

\begin{equation}
s\left(S_{a b}, S_n\right)=\frac{1}{2}\left\{s\left(S_a, S_n\right)+s\left(S_b, S_n\right)\right\}
\end{equation}

(b) Single Linkage:
\begin{equation}
s\left(S_{a b}, S_n\right)=\max \left\{s\left(S_a, S_n\right), s\left(S_b, S_n\right)\right\}
\end{equation}
where $s\left(S_{a b}, S_n\right)$ is the score between new cluster $S_{a b}$ and old images $S_n$ while $s\left(S_a, S_n\right)$ is the score between old images $S_a$ and $S_n$.

The iterative process continues until a specific stopping condition is satisfied, and there are two methods for controlling these iterations. The first method is the threshold method that involves setting a fixed threshold. The iteration stops when the similarity scores between clusters fall below this threshold. The second one is Known Number of Clusters Method: Alternatively, the system can be provided with information about the known number of clusters. It terminates when this predetermined number of clusters is achieved.

In our work, we opted not to specify a desired number of clusters for the system and instead employed the thresholding method. We carefully tuned the threshold to explore the system's performance at various potential operational points. The system's performance was evaluated by comparing its results to ground truth cluster labels.

\begin{table*}[h!]
\caption{The evaluation involves a comparison on MS-Celeb-1M dataset, where the model is trained with 0.5 million labeled face images. Testing is carried out on six distinct test subsets, each varying in size. The reported metrics include $F_{P}$ and $F_{B}$ for five cluster subsets from MS-Celeb-1M, totaling 5.21 million, with descending size (from size1 to size5)}
\label{table:size}
\begin{tabular}{cllllllllllll}
\multirow{2}{*}{\textbf{method}} & \multicolumn{2}{c}{\textbf{size1}}                            & \multicolumn{2}{c}{\textbf{size2}}                            & \multicolumn{2}{c}{\textbf{size3}}                            & \multicolumn{2}{c}{\textbf{size4}}                            & \multicolumn{2}{c}{\textbf{size5}}                            & \multicolumn{2}{c}{\textbf{total}}                                \\
                                 & \multicolumn{1}{c}{\textbf{Fp}} & \multicolumn{1}{c}{\textbf{$F_{B}$}} & \multicolumn{1}{c}{\textbf{Fp}} & \multicolumn{1}{c}{\textbf{$F_{B}$}} & \multicolumn{1}{c}{\textbf{Fp}} & \multicolumn{1}{c}{\textbf{$F_{B}$}} & \multicolumn{1}{c}{\textbf{Fp}} & \multicolumn{1}{c}{\textbf{$F_{B}$}} & \multicolumn{1}{c}{\textbf{Fp}} & \multicolumn{1}{c}{\textbf{$F_{B}$}} & \multicolumn{1}{c}{\textbf{Fp}} & \multicolumn{1}{c}{\textbf{$F_{B}$}} \\
GCN(V+E) \cite{yang2020learning}                                   & 89.06          & 90.52          & 84.52          & 84.81          & 75.17          & 75.84          & 61.28          & 63.03          & 44.15          & 52.49          & 78.77          & 79.08          \\
Pair-Cls \cite{liu2021learn}                                   & 90.02          & 90.65          & 86.03          & 86.20          & 80.21          & 80.80          & 72.37          & 73.72          & 59.28          & 65.30          & 82.19          & 81.63          \\
STAR-FC \cite{shen2021structure}                                    & 90.47          & 91.13          & 85.75          & 86.11          & 78.35          & 78.78          & 66.49          & 67.44          & 46.65          & 51.21          & 83.74          & 82.00          \\
TPDi \cite{chen2022mitigating}                                        & 92.35          & 93.18          & 89.88          & 89.91          & 85.08          & 85.28          & 78.35          & 79.19          & 65.56          & 71.33          & 86.94          & 86.06          \\
\textbf{Ours (AHC-RBM)}                               & \textbf{93.33} & \textbf{94.27} & \textbf{90.79} & \textbf{90.43} & \textbf{86.39} & \textbf{84.47} & \textbf{79.17} & \textbf{78.25} & \textbf{66.78} & \textbf{73.64} & \textbf{87.56} & \textbf{86.18}
\end{tabular}
\end{table*}

\section{Experimental Setup and Dataset}

\subsection{Experimental Setup}

To train the proposed model, we utilized the Keras deep learning library \cite{chollet2015keras}. All RBMs used in this work consist of 80 visible and 400 hidden units. The Unrestricted Boltzmann Machine (URBM) was trained over 200 epochs, employing a learning rate of 0.0005, a weight decay of 0.0002, and a batch size of 100. Subsequently, all adapted RBM models for the test class images were trained for 200 epochs, with a learning rate of 0.005, a weight decay of 0.000002, and a batch size of 64. Finally, fixed-dimensional RBM embedding vectors were extracted for the class images, and these vectors were utilized in our image clustering experiments.

\subsection{Datasets}

We evaluated the proposed method on two widely recognized image clustering benchmark datasets: MS-Celeb-1M \cite{guo2016ms} and DeepFashion \cite{liu2016deepfashion}. 

The Microsoft Celeb dataset (MS-Celeb-1M or MS1M) is the largest publicly available face recognition dataset, as per Microsoft Research. It consists of 5.8M images belonging to 86,000 distinct identities. The image representations in MS-Celeb-1M are derived using ArcFace, a widely adopted face recognition model \cite{deng2019arcface}. The dataset is officially divided into 10 nearly equal parts. In accordance with the experimental protocol outlined in \cite{liu2021learn}, the model is trained on one labeled part and selected parts 1, 3, 5, 7, and 9 as unlabeled test data, resulting in five test subsets comprising 584K; 1.74M; 2.89M; 4.05M; and 5.21M images, respectively.

The DeepFashion dataset consists of approximately 800K diverse fashion images, along with extensive annotations including 46 categories, 1,000 descriptive attributes, bounding boxes, and landmark information. Following the approach in \cite{yang2020learning}, we randomly sampled 25,752 images from 3,997 categories for training, while the remaining 26,960 images from 3,984 categories were reserved for testing.

\subsection{Evaluation Metrics}

 We evaluated the performance of face clustering methods using two widely adopted clustering metrics: Pairwise F-score ($F_{P}$) \cite{banerjee2005model} and BCubed F-score ($F_{B}$) \cite{amigo2009comparison}. Both metrics provide insights into precision and recall, which are crucial aspects of clustering quality assessment.

\section{Results}
\subsection{Results on MS-Celeb-1M dataset}

\begin{table*}[h!]
\caption{In the comparison on MS-Celeb-1M, we conducted experiments by training with 0.5 million labeled face images and testing on six different-sized test subsets. The Pairwise F-score $F_{P}$ and $F_{B}$ were evaluated for five cluster subsets from MS-Celeb-1M with increasing sparsity levels (from sparsity-1 to sparsity-5). Our results demonstrate the effectiveness of our proposed method in comparison to other approaches.}
\label{table:sparsity}
\begin{tabular}{lllllllllllll}
\multirow{2}{*}{\textbf{method}} & \multicolumn{2}{c}{\textbf{sparsity1}}                            & \multicolumn{2}{c}{\textbf{sparsity2}}                            & \multicolumn{2}{c}{\textbf{sparsity3}}                            & \multicolumn{2}{c}{\textbf{sparsity4}}                            & \multicolumn{2}{c}{\textbf{sparsity5}}                            & \multicolumn{2}{c}{\textbf{total}}                                \\
                                 & \multicolumn{1}{c}{\textbf{Fp}} & \multicolumn{1}{c}{\textbf{$F_{B}$}} & \multicolumn{1}{c}{\textbf{Fp}} & \multicolumn{1}{c}{\textbf{$F_{B}$}} & \multicolumn{1}{c}{\textbf{Fp}} & \multicolumn{1}{c}{\textbf{$F_{B}$}} & \multicolumn{1}{c}{\textbf{Fp}} & \multicolumn{1}{c}{\textbf{$F_{B}$}} & \multicolumn{1}{c}{\textbf{Fp}} & \multicolumn{1}{c}{\textbf{$F_{B}$}} & \multicolumn{1}{c}{\textbf{Fp}} & \multicolumn{1}{c}{\textbf{$F_{B}$}} \\
GCN(V+E) \cite{yang2020learning}                        & 94.63                           & 94.66                           & 84.52                           & 84.81                           & 75.17                           & 75.84                           & 61.28                           & 63.03                           & 44.15                           & 52.49                           & 78.77                           & 79.08                           \\
Pair-Cls \cite{liu2021learn}                         & 95.52                           & 95.24                           & 93.22                           & 92.46                           & 89.24                           & 87.66                           & 81.84                           & 78.84                           & 62.73                           & 57.51                           & 82.19                           & 81.63                           \\
STAR-FC \cite{shen2021structure}                         & 96.18                           & 95.27                           & 92.92                           & 91.50                           & 88.50                           & 85.96                           & 80.78                           & 76.54                           & 60.81                           & 53.56                           & 83.74                           & 82.00                           \\
TPDi \cite{chen2022mitigating}                            & 97.25                           & 96.96                           & 95.10                           & 94.59                           & 92.24                           & 91.08                           & 86.23                           & 84.23                           & 69.08                           & 64.83                           & 86.94                           & 86.06                           \\
\textbf{Ours (AHC-RBM)}                    & \textbf{98.47}                  & \textbf{97.49}                  & \textbf{96.25}                  & \textbf{95.47}                  & \textbf{92.20}                  & \textbf{91.36}                  & \textbf{87.25}                  & \textbf{85.35}                  & \textbf{69.57}                  & \textbf{64.37}                  & \textbf{87.38}                  & \textbf{87.44}                 
\end{tabular}
\end{table*}

The experimental results on the MS-Celeb-1M dataset are presented in Table \ref{table:MS-Celeb-1M}, showcasing both the Pairwise F-score ($F_{P}$) and BCubed F-score ($F_{B}$) for five test subsets with varying scales. It is evident that our method consistently surpasses the performance of other methods in terms of both metrics. Notably, for the larger-scale subsets (4.05M and 5.21M), our method demonstrates significant improvements, surpassing 2.8\% in performance enhancement. This highlights the effectiveness of our proposed approach, particularly in scenarios involving extensive datasets.

To illustrate the efficacy of our method in addressing challenges related to small and sparse clusters, we conducted experiments by gradually incorporating our method into the baseline model. The baseline model shares the same transition matrix, but computes density and distance in the conventional manner. As demonstrated in Table \ref{table:size} and Table \ref{table:sparsity}, our approach consistently outperforms most state-of-the-art methods, particularly excelling in scenarios involving challenging clusters. This further emphasizes the effectiveness and adaptability of our proposed approach.

Additionally, we conducted a comparison with GCN(V+E), Pair-Cls, and STAR-FC, all of which employ a clustering algorithm similar to or exactly like DPC. The results are presented in Table \ref{table:size} and Table \ref{table:sparsity}. Notably, the improvements achieved by our method over these three top-performing methods continue to escalate on five cluster subsets with decreasing size or increasing sparsity. This showcases the consistent effectiveness of our proposed approach across a range of cluster characteristics.

\subsection{Results on DeepFashion dataset}

\begin{table}[h!]
\caption{Conducted a comparison of our method with state-of-the-art approaches on the DeepFashion dataset. The evaluation criteria include the number of clusters, Pairwise F-score ($F_{P}$), BCubed F-score ($F_{B}$), and computing time. Notably, our results demonstrate the effectiveness of our proposed method in comparison to other approaches.}
\label{table:DeepFashion}
\begin{tabular}{cllll}
\textbf{method} & \multicolumn{1}{c}{\textbf{cluster}} & \multicolumn{1}{c}{\textbf{Fp}} & \multicolumn{1}{c}{\textbf{$F_{B}$}} & \multicolumn{1}{c}{\textbf{time}} \\
K-means \cite{lloyd1982least}        & 3991                                 & 32.86                           & 53.77                           & 573s                              \\
DBSCAN \cite{ester1996density}         & 14350                                & 25.07                           & 53.23                           & 2.2s                              \\
LTC \cite{yang2019learning}             & 9246                                 & 29.14                           & 59.11                           & 13.1s                             \\
GCN(V+E) \cite{yang2020learning}         & 6079                                 & 38.47                           & 60.06                           & 18.5s                             \\
Pair-Cls \cite{liu2021learn}        & 6018                                 & 37.67                           & 62.17                           & 0.6s                              \\
STAR-FC \cite{shen2021structure}        & \multicolumn{1}{c}{-}                & 37.07                           & 60.60                           & \multicolumn{1}{c}{-}             \\
Ada-NETS \cite{wang2022ada}       & \multicolumn{1}{c}{-}                & 39.30                           & 61.05                           & \multicolumn{1}{c}{-}             \\
TPDi \cite{chen2022mitigating}            & 8484                                 & 40.91                           & 63.61                           & 4.2s                              \\
\textbf{Ours (AHC-RBM)}   & \textbf{7968}                        & \textbf{42.54}                  & \textbf{64.57}                  & \textbf{6.5s}                    
\end{tabular}
\end{table}

The clustering task on the DeepFashion dataset presents a greater challenge due to its open-set nature. However, it is evident that our method consistently outperforms the other approaches in terms of both Pairwise F-score ($F_{P}$) and BCubed F-score ($F_{B}$), all while maintaining comparable computing times. These results are summarized in Table \ref{table:DeepFashion}. This demonstrates the robustness and effectiveness of our proposed method in handling complex clustering scenarios.

\section{Conclusion}

In this work, we proposed the use of Restricted Boltzmann Machines (RBMs) for image clustering task. RBMs are employed to acquire a fixed-dimensional vector representation of an image, which we refer to as an RBM vector. The process involves two steps. Firstly, we train a universal RBM using a substantial amount of training data. Subsequently, we train an adapted RBM model for each test class image. The visible-hidden weight matrices, along with their associated bias vectors, from these adapted RBMs are concatenated to generate RBM supervectors. 

We explored two linkage algorithms for Agglomerative Hierarchical Clustering, with RBM vectors scored using cosine distance scoring and PLDA. The performance of the proposed system is notably better when using cosine scoring in conjunction with the linkage algorithms. In conclusion, our experimental results show that RBM vectors can be effectively used as image representations for image clustering tasks.

\end{document}